# Context-aware Sentiment Word Identification: sentiword2vec

Yushi Yao, Guangjian Li

Department of Information Management, Peking University, China

{yaoyushi, ligj}@pku.edu.cn

**Abstract**

Traditional sentiment analysis often uses sentiment dictionary to extract sentiment information in text and classify documents. However, emerging informal words and phrases in user generated content call for analysis aware to the context. Usually, they have special meanings in a particular context. Because of its great performance in representing inter-word relation, we use sentiment word vectors to identify the special words. Based on the distributed language model word2vec, in this paper we represent a novel method about sentiment representation of word under particular context, to be detailed, to identify the words with abnormal sentiment polarity in long answers. Result shows the improved model shows better performance in representing the words with special meaning, while keep doing well in representing special idiomatic pattern. Finally, we will discuss the meaning of vectors representing in the field of sentiment, which may be different from general object-based conditions.

## 1 Introduction

Different from traditional news corpus, user generated content is informal in linguistics. While they constantly create new words and phrases, some informal words express more meaning. It challenges traditional sentiment analysis methods as provide us more vivid corpus to explore human beings' sentiment expression.

Exploring this kind of new words requires deeply making use of context, especially semantic meaning. Using distributed language methods in a particular context provide insight to latent language meaning, while shows superiority in context aware and analogy.

Sometimes, in a special language context, some words would have special meaning different from normal environment. As the social media developing, more online community the cluster effect. For example, "refugee" is a positive word in the discussion of human rights union while negative among the real estate holder.

In this paper, we use a model based on word2vec to find out the special word in a particular context. Different from researches on short information flow like twitter, we use long articles, answers to a question posted in social media as corpus. Using our model, we provide a method dipping into the latent sentiment tendency in long social articles. After training vectors using word2vec, we change the vectors of words with known sentiment polarity and train them again controlling iteration times. The special words in a particular context are detected in the model, while a better vector expression of them is presented.

## 2 Literature Review

As sentiment reflects more latent information in text, the meanings that sentiment words contain are often context-specific. This nature led to general sentiment feature processing problematic. Common is that a word, or phrase, is positive in one context while negative in another. So considering more contextual-information such as topic or domain is essential. In this process, identifying new sentiment words and new meaning of sentiment words in context is important. Additionally, informal words and phrases play an important role in user generated content (UGC). But the emerging informal words usually do not disappear in sentiment dictionaries. So to address the special words is a key issue.

Though studies exploring sentiment features are plentiful, many of which is based on N-gram (Cambria, Havasi & Hussain, 2012). In other words, the construction of sentiment features usually shows have dependency on context. However, it is not enough. The latent nature of sentiment word meaning led to special processing in identification.

Researchers have spent a lot of time on finding and selecting typical sentiment features. However, this traditional way to construct sentiment features makes it difficult to go further in latent sentiment features (Asghar, Khan, Ahmad & Kundi, 2014), especially context-aware information.

## 2.1 Lexicon-based Identification

In a particular context, the sentiment polarity of some words may differ from its polarity in general context. In forum discussion, an unknown file is neural, while an unknown politician is usually negative. In a way, politician is regarded negative in this context. It is difficult for normal sentiment analysis to recognize this special sentiment expression. So, we need special processing to recognize and predict the rich meaning those words contain.

Recognizing such special lexicon first is a common method. A traditional approach starts from semantic lexicon such as WordNet to build a conceptual semantic dictionary such as WordNet-Affect (Strapparava & Valitutti, 2004), SenticNet (Cambriaet al, 2012) etc. Common further approach is to identify words that might have special sentiment polarity in specific situations first. Out of this kind of lexical ambiguity, Weichselbraun, Gindl and Scharl (2013) recognized ambiguous sentiment terms from corpus and predict these special terms according to context. Similar technique was used in research of Jijkoun, Rijke and Weerkamp (2010), while ambiguous sentiment terms were replaced by target entities in reviews. Not satisfied with domain-specific sentiment lexicons, Lu, Castellanos, Dayal and Zhai (2011) paid further attention to sentiment polarity of the target entities, especially how a word performs different meaning in different aspects.

All of these start from sentiment terms, doing recognition, processing, and then sentiment analysis. The fine processing performs well in sentences while limited by the accuracy of concept recognition. Additionally, focusing on target entities in reviews limits its generalization in more complex context.

## 2.2 Context-based Identification

From another perspective, recognizing what the context is also helpful.

Wilson, Wiebe, & Hoffmann (2005) started from the polarity change of phrase-level clue by recognizing the shifter words which change sentiment polarity. However, this fully supervised method is of high cost, while the shifter words are able to reflect more information under different context. Kanayama & Nasukawa (2006), on the other hand, used statistical methods to determine correct sentiment word candidates. Their work is limited in sensitivity to wider context.

So, in the process of correcting sentiment lexicon, researchers try to make use of more context information. Qiu, Liu, Bu, & Chen (2009) used dependency tree to extract new sentiment words in diverse product domains, while Lu et al. (2011) focused on different aspects of the same domain. Choi, Kim & Myaeng (2009) went further by using sentiment clues to identify domain-context features. As most researchers do, they used seed words to find more sentiment words, they use limited sentiment clues and bootstrap algorithm to converge and aggregate new clues. Such topic-context features excluded from domain-specific work well in filed-related query. However, language in the news corpus they used is rather formal, while nowadays informal language acts an increasingly important role in user generated content.

More context information is waiting for use. Aisopos, Papadakis, Tserpes & Varvarigou (2012) tried to use context technique in analyzing microblogs. The structure of social network provided them plentiful information about context, so they used multiple features about authors, followers, friends, in a word, what a group of people a tweet is in.

## 2.3 Vectors in Identification

Distributed language provides a different and flexible insight into these words' semantic relation. As a result, it attracts researchers' attention in applying to sentiment analysis.

Some focus on making use of the special relation of word vectors, while others paid attention to sentiment vectors. Xia, Wang, Dai and Li (2013) used original and antonymous views in pairs in classifier training, which gives full play word vectors' advantage in latent meaning.

Others would like to improve sentiment words' vector representation. Tang et al. (2014) use a sentiment node in the C&W model, in order to get sentiment feature and general linguistic feature at the same time. Also, another two-dimension softmax layer were added to get words' sentiment vector in SSWE model.

These models provide good ways to get vector containing sentiment information in short messages. However, their methods require high-cost dataset, or did not show whether their sentient vectors are sensible to context. In this paper, we would provide a method to address the special sentiment words and see how the context change words' sentiment polarity.

## 3 Model

As the special words we focusing on in this paper are actually components of a kind of abnormal concept construction. Where there are concepts with special meaning in a context, there exist words with abnormal sentiment. Long articles with personal opinions and sentiments are in need for developing such special words. Rather than micro-blogs such as twitter, which have been widely studied, the long answers, especially the online communities talking about society and personal experience, are more likely to construct typical context-aware concepts and corresponding sentiment abnormality. In this respect, we choose the most popular online Q&A community, Zhihu as our research object, where participants talking about their personal experiences and opinions.

In this section, we will talk about our method. We talk about how to process words first and then go to the way to train them and get vectors. Then, the evaluation methods will be discussed.

### 3.1 Seed word

As most sentiment word identification methods do, we use a set of seed sentiment words in our method. Because of its high cost, there is little sentiment dictionary describing whether a word is more positive than another word. So, at the beginning, limited by sentiment dictionaries, we donate the vectors of all positive words as <1, 1, 1, … 1>, which means all of these seed words have the same meaning in the space. Of course it is beyond the truth. In later process, we will give them more abundant meaning.

## 3.2  General Vector

Similar with most model, Mikolov's (2013) model carries the entity and topic information in features. Each word has a n-dimension vector to represent its meaning in the language space. In the skip-gram model, a target word is set at the input layer, while the context words are on the output layer. Then we have a hidden-layer, whose output h is simply copying (and transposing) a row of the input multiply hidden weight matrix, W, associated with the input word. On the output layer, the nearby words' vectors would be updated according to the hidden layer.

In our experiment, we use the Wikipedia corpus to pre-train and get the word vectors first. These vectors would show relationship between words in general context. They help to describe general corpus and avoid overfitting in a particular context. Then we continued using the Zhihu corpus to train the context-related word vectors.

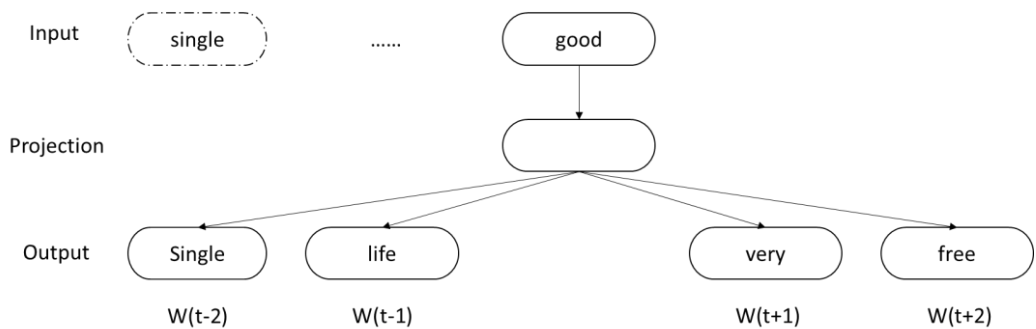

## 3.3  Sentiment Vector

However, sentiment information is hidden deeper. In order to express more sentiment information in feature and find out the special words in a particular context, we add a vector update training using sentiment dictionary.

After using a field-related corpus to train a word2vec model and get the general vector, we use the sentiment dictionary to update the sentiment words' vector as the following chart shows. Positive words' vectors are set <1, 1…>, and negative ones are<-1, -1…>. Now the sentiment words' vector could easily show their sentiment polarity while others do not. Next, we begin another iteration to update the words' vectors. Words near a lot of positive sentiment words will become positive. How positive a word is determined by its surroundings, in other words, its nearby sentiment words. In this step, a word's sentiment polarity is influenced by both its context and original meaning. The special words, which appear many times, would has stronger weight on context sentiment.

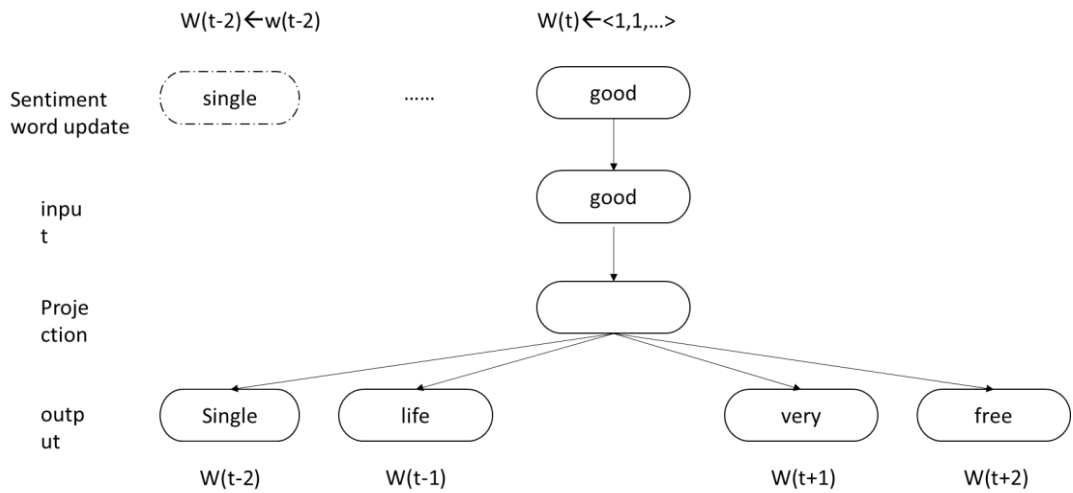

Picture1 our model based on word2vec

## 3.4 Sentiment Polarity

In distributed language model, a word is defined by its vector and distance with other words. To define a word's sentiment polarity, we select a set of typical positive words and negative words, and define the mean of positive set's vector as the positive vector, so as the negative one. So, once we know a word's distance between positive vector and negative vector, we know its sentiment tendency.

To compare the ability word vectors express context-related words among the three models, we selected X words from test corpus, and compare the distance between positive words and negative words. If a word is more distant from positive words than negative words, then it is negative. In this way, we decided whether a word is positive or negative.

Given precision as the number of correctly recognized words dividing the amount, we will calculate the precision of the vectors after training.

## 3.5 Classifier

After getting the sentiment word vectors, we use the stable classifier, support vector machine, to classify the answers, to see the performance of different vectors. The answers were labeled positive or negative. Features are the vectors we got in preceding steps. To avoid interference, we use the simple SVM and popular RBF function as kernel function. In our experiment, we marked about 4000 pieces of answers from Zhihu corpus. Using 80% for classifier training and 20% for predicting. Word vectors from Wikipedia training, Zhihu corpus training and Zhihu sentiment training were respectively used for classification.

# 4   Experiment

In this section, we will talk about the process of our experiment. First, we will discuss the way to get sentiment dictionary and to deal with them. Then, we will process the corpus to get it well-prepared. Next, we will train them and get the vectors. Finally, we will compare the performances of the three set of vectors on both abnormal sentiment word classification and classification of the whole answers.

## 4.1   Sentiment Dictionary

Limited by lack of dictionaries marked words' Intensity of sentiment, we only use sentiment tendency in this model.

In the field of Chinese sentiment, though the research of sentiment classification has been plentiful over decades, a comprehensive sentiment dictionary is in lack. To make the dictionary we used mature, we use the widely-used Hownet[1] and NTUSD (Ku & Chen, 2007) as the base of our sentiment dictionary. Opencc[2] is used to translate traditional Chinese in NTUSD to simplified Chinese. Then, we merged the words having same tendency in both dictionary and removed the ones having different tendency. The sentiment dictionary we used in experiment has more than 18000 words.

## 4.2   Experiment Preparation

Existing sentiment corpus are extract from news or short message from social media lie twitter. However, language in the former is quite formal, while the latter only contains short messages providing limited context feature. So, both of them are not appropriate for our experiments. So we use a new corpus extracting from internet.

Corpus preparation

To compare vectors in particular context and general context, we crawlled about 500M articles from Chinese Wikipedia for comparison experiment.

To get typical corpus in the context Zhihu, we selected three characteristic topics: "How to evaluate X", "Individual consulting", and "What is the experience of X". All of these three topics have more than 1000 star questions, which is the upper bound that Zhihu shows. Additionally, they are more subjective and relate closely to personal experience. So, they would use more context-related words rather than science topics. Especially, to make sure the answers are typical, we selected the top 20 answers of each question. Below is a brief description of the corpus in our experiment.

---

[1] http://www.keenage.com/

[2] http://opencc.byvoid.com/

Table1Corpus Description

| Topics | Messages |
|---|---|
| How to evaluate X | 17874 |
| Individual Consulting | 19419 |
| What is the experience of X | 23429 |
| Total Amount | 60722 |

Preprocess

The language in Wikipedia and Zhihu is various. So we delete Japanese, Korean etc. in corpus and translate traditional Chinese to simplified Chinese using opencc[3] to get rid of interference.

In need of recognize new words and special words in this context, we need tokenizer with larger granularity. Jieba[4] is used to tokenize the corpus and we get more than 210000 different words. About 1800 stop words were deleted in the training corpus.

## 4.3 Experiment Analysis

In following analysis, we would compare the performance of our model on Zhihu corpus, word2vec on Zhihu corpus and word2vec on Wikipedia corpus. Results show that our model is more accurate when recognizing the words with special meaning in a particular context.

### 4.3.1 Words Representation

As stated before, some words may have different sentiment polarity in different context. Below are examples of words in different context：

Table2Words in different context

| Word sample | General context | Zhihu context |
|---|---|---|
| Test paper | ↓ | ↑ |
| Parade | ↓ | ↑ |
| Single | ↓ | ↑ |

---

[3]https://github.com/BYVoid/OpenCC

[4]https://github.com/fxsjy/jieba

We selected the words appeared more than 5 times in corpus for training. After exclude stop words, we analyzed about 100 typical words' sentiment tendency. The three models perform differently as the following table shows.

Table3Sentiment word prediction accuracy

|  | Word2vec in Wikipedia | Word2vec in Zhihu | Our model in Zhihu |
|---|---|---|---|
| Prediction | 55% | 62% | 65% |

Generally speaking, when a word does not have special meaning in the context, the three models' vector give similar sentiment tendency. For example, "expectation" "quality"" startup" are positive, while "complex" and " drug" are negative。

Training on Zhihu corpus, both word2vec model and our model shows better performance on sentiment word tendency. It shows the sentiment tendency of words out of language using custom could be improved by using appropriate corpus.

However, some words change. On one hand, some words have similar meaning, but the objects referred have different reputation in different context. On the other hand, the words have completely different meaning in a new context.

Table4words in different context・group one

|  | model | | |
|---|---|---|---|
| Word sample | Word2vec in Wikipedia | Word2vec in Zhihu | Our model in Zhihu |
| 1 |  |  |  |
| Single | ↓ | ↑ | ↑ |
| hoodwink | ↓ | ↑ | ↑ |
| Shy | ↑ | ↓ | ↓ |
| Bottom | ↑ | ↓ | ↓ |
| 2 |  |  |  |
| chrysanthemum[5] | ↑ | ↓ | ↓ |

---

[5] Its original meaning is a well-known flower, while extended meaning on Internet is anal, a sexual metaphor.

However, when we focus on the words carrying more opinion polarity, changes occur. Our model performed better in finding them out and point out their sentiment polarity. Our model strengthens their sentiment information in vectors. The difference may result from interference of words' thematic meaning, which leads to insensitivity to sentiment.

Table5 words in different context・group two

|  | Model | | |
|---|---|---|---|
| Word sample | Word2vec in Wikipedia | Word2vec in Zhihu | Our model in Zhihu |
| 3 | | | |
| feminist | ↑ | -- | ↑ |
| State administration of press, publication, radio, film and television[6] | ↑ | ↑ | ↓ |
| Yimou Zhang[7] | ↑ | ↑ | ↓ |
| Jingming Guo[8] | ↑ | -- | ↓ |
| Idealistic | ↑ | -- | ↓ |
| The second generation[9] | ↑ | -- | ↓ |
| caprice | ↓ | ↓ | ↑ |

Notably, the sentiment carried in word vector comes from user generated corpus, so the sentiment is subjective. However, subjective feeling does not equal to sentiment to the corresponding object. For example, a negative word may suggest a bad one, or the user feeling disappointed because of limited ability. Such a case is the series of math courses. Surely the word sentiment could not represent users' opinion completely, more information about context is in need.

Table6 words in different context・group three

---

[6] The bureau' rating system spark dissatisfaction among some people.

[7] Acontroversial film director.

[8] Ayouth literature author, embroiled in controversy because of plagiarizing.

[9] The son or daughter of the officers, the rich, the founders of the nation.

|  | Model | | |
|---|---|---|---|
| Word sample | Word2vec in Wikipedia | Word2vec in Zhihu | Our model in Zhihu |
| 4 | | | |
| Mathematical analysis | ↑ | -- | ↓ |
| Math problems | ↑ | -- | ↓ |

As the following figure[10] shows, most words predicted negative in model three while positive in first two models are closely related to reality, such as Department of the Communist, State administration of press, publication, radio, film and television etc. in the contrast, words positive in model three and negative in the other two models are words closely related to life and personal interests, such as cat, third element[11] etc.

---

[10] The words showed in this figure have special meaning in this context. For example, "idea" means creative in general context while means lowpower of execution in Zhihu. "Stubborn" means caprice in general content while means unremitting effort in Zhihu. The words having special meaning have different location in the three models.

[11] Correspondingto second element (comics world), meaning real world.

Figure2 Word Vectors; Cross-model Comparison

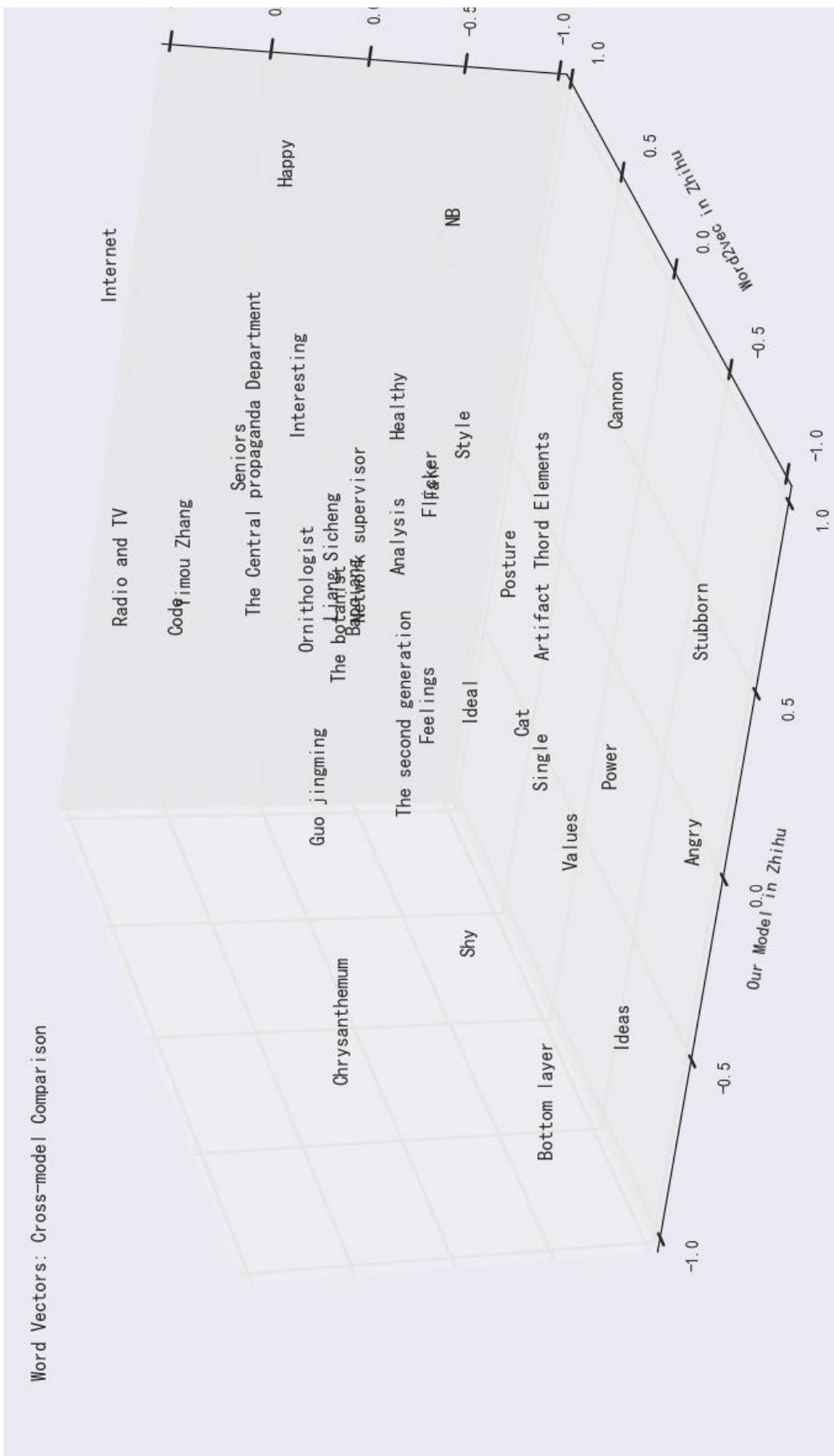

**4.3.2 Sentiment Classification**

While identify sentiment words, we also test the performance applying word vectors into sentiment classification task. In our experiment, we applied the feature vectors we get in new model into the classical binary classification task, so as to see how the vectors work in classification.

To show the performance of different feature vectors, we use the wide-spread, stable classification model SVM to do the classification task. We randomly selected about 9000 articles and tagged their sentiment. 80% of them were randomly selected as training set and the remaining as test set. Weighted feature vectors from the three models were separately used in training and predicting. We use RBF function as kernel function and set gamma as 0.7. Following chart shows the performance.

Table7sentiment classification prediction

| Vector model | Corpus | Sentiment dictionary | prediction |
| --- | --- | --- | --- |
| word2vec | Wikipedia | No | 67.61% |
| word2vec | Zhihu | No | 69.82% |
| Our model | Zhihu | Hownet+NTUSD | 70.00% |

The prediction result shows context-related training corpus has better performance. However, because of limited "feature words" in the whole articles, our model does not outperform other models significantly. It seems a targeted model is in need to help the featured words to play better in classification.

# 5 Discussion & Conclusion

Generally, each column of vectors in word2vec is thought to be representing meaning from different aspects. However, Tang et al. (2014) tried to produce a whole vector representing sentiment based on that. That leads to a question: could the vectors represent different aspects of sentiment? For example, do you agree with that topic? Do you think it fair? Does it benefit you? The experiment stated in this paper is an exploration. Our experiment shows that in the field of sentiment, the vectors may represent sentiment meanings rather than different aspects of objects. However, it remains a guess. Further research is in demand to know the meaning of the vectors.

In this paper, we provided a new way to predict the special words' sentiment polarity. These special words are redefined in a particular context. They have different meaning from what they are in general context while represent the opinion of users in the particular context, community, forum etc. However, corpus in a particular context is limited. The robustness and ability of generalization still need verification.

# 6 Reference

Aisopos, F., Papadakis, G., Tserpes, K., & Varvarigou, T. (2012, June). Content vs. context for sentiment analysis: a comparative analysis over microblogs. In *Proceedings of the 23rd ACM conference on Hypertext and social media* (pp. 187-196). ACM.

Asghar, M. Z., Khan, A., Ahmad, S., & Kundi, F. M. (2014). A review of feature extraction in sentiment analysis. *Journal of Basic and Applied Scientific Research*, *4*(3), 181-186.

Cambria, E., Havasi, C., & Hussain, A. (2012, May). SenticNet 2: A Semantic and Affective Resource for Opinion Mining and Sentiment Analysis. In *FLAIRS conference* (pp. 202-207).

Choi, Y., Kim, Y., & Myaeng, S. H. (2009, November). Domain-specific sentiment analysis using contextual feature generation. In *Proceedings of the 1st international CIKM workshop on Topic-sentiment analysis for mass opinion* (pp. 37-44). ACM.

Jijkoun, V., de Rijke, M., & Weerkamp, W. (2010, July). Generating focused topic-specific sentiment lexicons. In *Proceedings of the 48th Annual Meeting of the Association for Computational Linguistics* (pp. 585-594). Association for Computational Linguistics.

Kanayama, H., & Nasukawa, T. (2006, July). Fully automatic lexicon expansion for domain-oriented sentiment analysis. In *Proceedings of the 2006 conference on empirical methods in natural language processing* (pp. 355-363). Association for Computational Linguistics.

Ku, L. W., & Chen, H. H. (2007). Mining opinions from the Web: Beyond relevance retrieval. *Journal of the American Society for Information Science and Technology*, *58*(12), 1838-1850.

Lu, Y., Castellanos, M., Dayal, U., & Zhai, C. (2011, March). Automatic construction of a context-aware sentiment lexicon: an optimization approach. In *Proceedings of the 20th international conference on World wide web* (pp. 347-356). ACM.

Mikolov, T., Chen, K., Corrado, G., & Dean, J. (2013). Efficient estimation of word representations in vector space. *arXiv preprint arXiv:1301.3781*.

Qiu, G., Liu, B., Bu, J., & Chen, C. (2009, July). Expanding Domain Sentiment Lexicon through Double Propagation. In *IJCAI* (Vol. 9, pp. 1199-1204).

Strapparava, C., & Valitutti, A. (2004, May). WordNet Affect: an Affective Extension of WordNet. In *LREC* (Vol. 4, pp. 1083-1086).

Tang, D., Wei, F., Yang, N., Zhou, M., Liu, T., & Qin, B. (2014, June). Learning Sentiment-Specific Word Embedding for Twitter Sentiment Classification. In *ACL (1)* (pp. 1555-1565).

Xia, R., Wang, C., Dai, X., & Li, T. Co-training for Semi-supervised Sentiment Classification Based on Dual-view Bags-of-words Representation.

Weichselbraun, A., Gindl, S., & Scharl, A. (2013). Extracting and grounding context-aware sentiment lexicons. *IEEE Intelligent Systems*, *28*(2), 39-46.

Wilson, T., Wiebe, J., & Hoffmann, P. (2005, October). Recognizing contextual polarity in phrase-level sentiment analysis. In *Proceedings of the conference on human language technology and empirical methods in natural language processing* (pp. 347-354). Association for Computational Linguistics.